\documentclass[10pt,twocolumn,letterpaper]{article}

\usepackage[pagenumbers]{iccv} % To force page numbers, e.g. for an arXiv version
\usepackage{xparse}

\usepackage{amsmath, amssymb, algorithm, algpseudocode, graphicx, booktabs}
\definecolor{iccvblue}{rgb}{0.21,0.49,0.74}
\usepackage[pagebackref,breaklinks,colorlinks,allcolors=iccvblue]{hyperref}

\def\paperID{*****} % *** Enter the Paper ID here
\def\confName{Preprint}
\def\confYear{2025}

\newcommand{\dataset}{WideRange4D\xspace}
\newcommand{\method}{Progress4D\xspace}

\title{\dataset: Enabling High-Quality 4D Reconstruction with \\Wide-Range Movements and Scenes}

\author{
\textbf{Ling Yang{$^{1,2  \dag}$}\thanks{Contributed equally.}, \ Kaixin Zhu{$^{1}$}\footnotemark[1], \ Juanxi Tian{$^{1}$}\footnotemark[1], \ Bohan Zeng{$^{1}$}\footnotemark[1]}\\ \textbf{Mingbao Lin{$^2$}, \ Hongjuan Pei{$^3$}, \ Wentao Zhang{$^{1}$}\thanks{This work was done when Ling Yang was an intern at Skywork AI. Contact: yangling0818@163.com, wentao.zhang@pku.edu.cn, yansc@nus.edu.sg}, \ Shuicheng Yan{$^{2,4 \dag}$}} \\ 
{$^1$}Peking University \quad {$^2$}  Skywork AI\\\quad {$^3$}University of the Chinese Academy of Sciences \quad {$^4$}National University of Singapore \\
\url{https://github.com/Gen-Verse/WideRange4D}
}

\begin{document}
\maketitle
\begin{abstract}
With the rapid development of 3D reconstruction technology, research in 4D reconstruction is also advancing, existing 4D reconstruction methods can generate high-quality 4D scenes. However, due to the challenges in acquiring multi-view video data, the current 4D reconstruction benchmarks mainly display actions performed in place, such as dancing, within limited scenarios. In practical scenarios, many scenes involve wide-range spatial movements, highlighting the limitations of existing 4D reconstruction datasets. Additionally, existing 4D reconstruction methods rely on deformation fields to estimate the dynamics of 3D objects, but deformation fields struggle with wide-range spatial movements, which limits the ability to achieve high-quality 4D scene reconstruction with wide-range spatial movements. In this paper, we focus on 4D scene reconstruction with significant object spatial movements and propose a novel 4D reconstruction benchmark, \dataset. This benchmark includes rich 4D scene data with large spatial variations, allowing for a more comprehensive evaluation of the generation capabilities of 4D generation methods. Furthermore, we introduce a new 4D reconstruction method, \method, which generates stable and high-quality 4D results across various complex 4D scene reconstruction tasks. We conduct both quantitative and qualitative comparison experiments on \dataset, showing that our \method outperforms existing state-of-the-art 4D reconstruction methods.
\vspace{-3mm}
\end{abstract}    
\section{Introduction}
\label{sec:intro}

% \begin{figure*}
%   \centering
%   \begin{subfigure}{0.68\linewidth}
%     \fbox{\rule{0pt}{2in} \rule{.9\linewidth}{0pt}}
%     \caption{An example of a subfigure.}
%     \label{fig:short-a}
%   \end{subfigure}
%   \hfill
%   \begin{subfigure}{0.28\linewidth}
%     \fbox{\rule{0pt}{2in} \rule{.9\linewidth}{0pt}}
%     \caption{Another example of a subfigure.}
%     \label{fig:short-b}
%   \end{subfigure}
%   \caption{Example of a short caption, which should be centered.}
%   \label{fig:short}
% \end{figure*}

\begin{figure}
    \centering
    \includegraphics[width=0.95\linewidth]{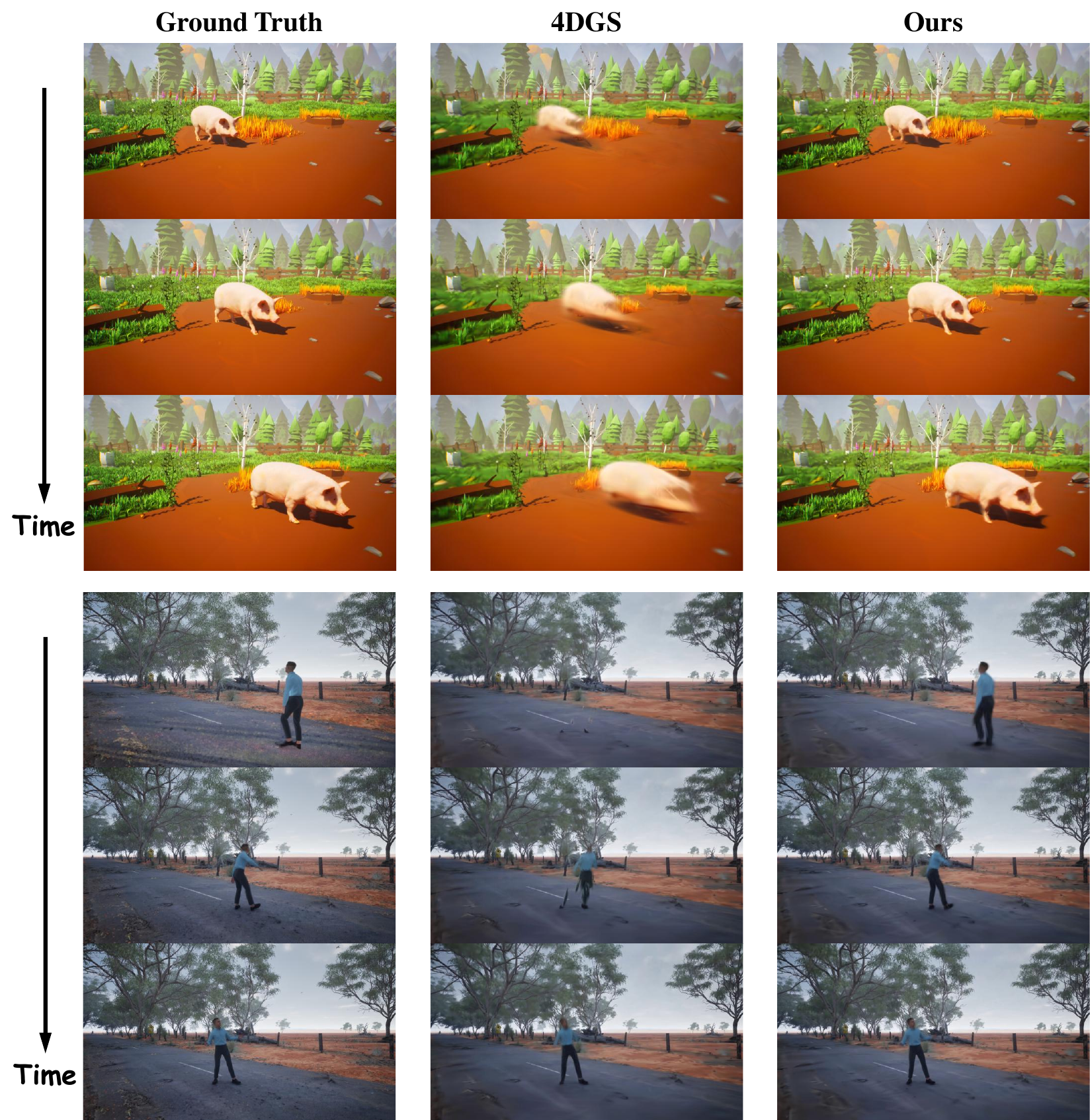}
    \caption{Visualization of the 4D scenes generated by our \method and 4DGS \cite{wu20244d} on the \dataset. Existing 4D reconstruction methods struggle to generate 4D scenes with wide-range movements, highlighting the need for our proposed new benchmark \dataset, and the high-quality 4D scenes produced by \method validate the effectiveness of our method.}
    \label{fig:first_comp}
    \vspace{-2mm}
\end{figure}

Constructing high-quality 4D scenes holds immense value for both academic research and industrial applications, and this task is highly challenging. With the continuous development of 3D generation technologies, some 4D reconstruction methods can achieve high-quality 4D generation in certain scenarios, positively contributing to the development of multimedia fields such as gaming and video.

Compared to the challenging task of acquiring 3D data, obtaining 4D data is more difficult. It requires capturing dynamic videos of scenes from different viewpoints within the same time period. Due to the challenges in data acquisition, particularly the difficulty of recording motion videos of the same scene from multiple cameras, each capturing different viewpoints, in real-world environments, existing 4D reconstruction methods often provide 4D data \cite{li2022neural, lin2022efficient, xu20244k4d} with only limited, localized motions, such as dancing in place, cooking, or assembling a bicycle, lacking 4D data involving wide-range spatial movements of objects in the scene.

Due to the limitations of existing 4D reconstruction benchmark, although 4D reconstruction methods \cite{xu20244k4d, wu20244d} can construct high-quality 4D results, they cannot verify the performance of these methods in 4D scenes involving wide-range spatial movements due to the limitations of test data. Furthermore, these methods utilize deformation fields to achieve dynamic 4D scenes, but the deformation computed by neural networks is suitable for actions performed in place. Existing 4D reconstruction methods generate 4D results with poor quality and blur when wide-range spatial movements of entire objects occur, as shown in Fig.~\ref{fig:first_comp}. Additionally, some video-to-4D generation methods \cite{wu2024sc4d, chu2024dreamscene4d} use segmentation and tracking models to generate 4D scenes with wide-range dynamics, however, the results generated by these methods generally have low quality.

In this paper, we focus on high-quality 4D scene reconstruction in complex scenarios. We curate a new test benchmark, \dataset, which contains various real and virtual scenes, each with an unspecified number of foreground subjects. The movement distance and action complexity of each foreground subject are diverse. Compared to previous 4D benchmark, \textbf{our \dataset greatly enhances the richness, diversity, and difficulty, providing a more comprehensive and effective evaluation of 4D generation methods}, as shown in Fig.~\ref{fig:teaser}. To achieve stable and high-quality 4D reconstruction, we propose \method, which divides the 4D generation process into two stages: high-quality 3D scene reconstruction and progressive fitting of 4D dynamics. This method ensures high-quality 4D scene reconstruction while also maintaining stability, especially in complex 4D scenes involving wide-range spatial movements. We validate the reconstruction ability of \method on our proposed \dataset, and experimental results show that \method outperforms existing state-of-the-art methods. Our contributions are as follows:
\begin{itemize}
    \item We introduce \dataset, a 4D benchmark that includes 4D scene data with wide-range spatial movements. Compared to existing 4D benchmarks, our dataset's richness and reconstruction difficulty are greatly enhanced.

    \item We propose a new 4D reconstruction method, \method. This method divides the 4D scene reconstruction process into two stages: high-quality 3D reconstruction and progressive fitting of 4D dynamics. Compared to existing 4D reconstruction methods, \method can more stably generate high-quality and rational 4D scenes.

    \item We conduct qualitative and quantitative comparisons on our proposed \dataset. The comparison results indicate that our \method achieves SOTA performance in reconstructing 4D scenes with wide-range spatial movements.

\end{itemize}

\section{Related Work}
\label{sec:relat}

\begin{figure*}
    \centering
    % \fbox{\rule{0pt}{6.5in} \rule{0.98\linewidth}{0pt}}
    \includegraphics[width=\linewidth]{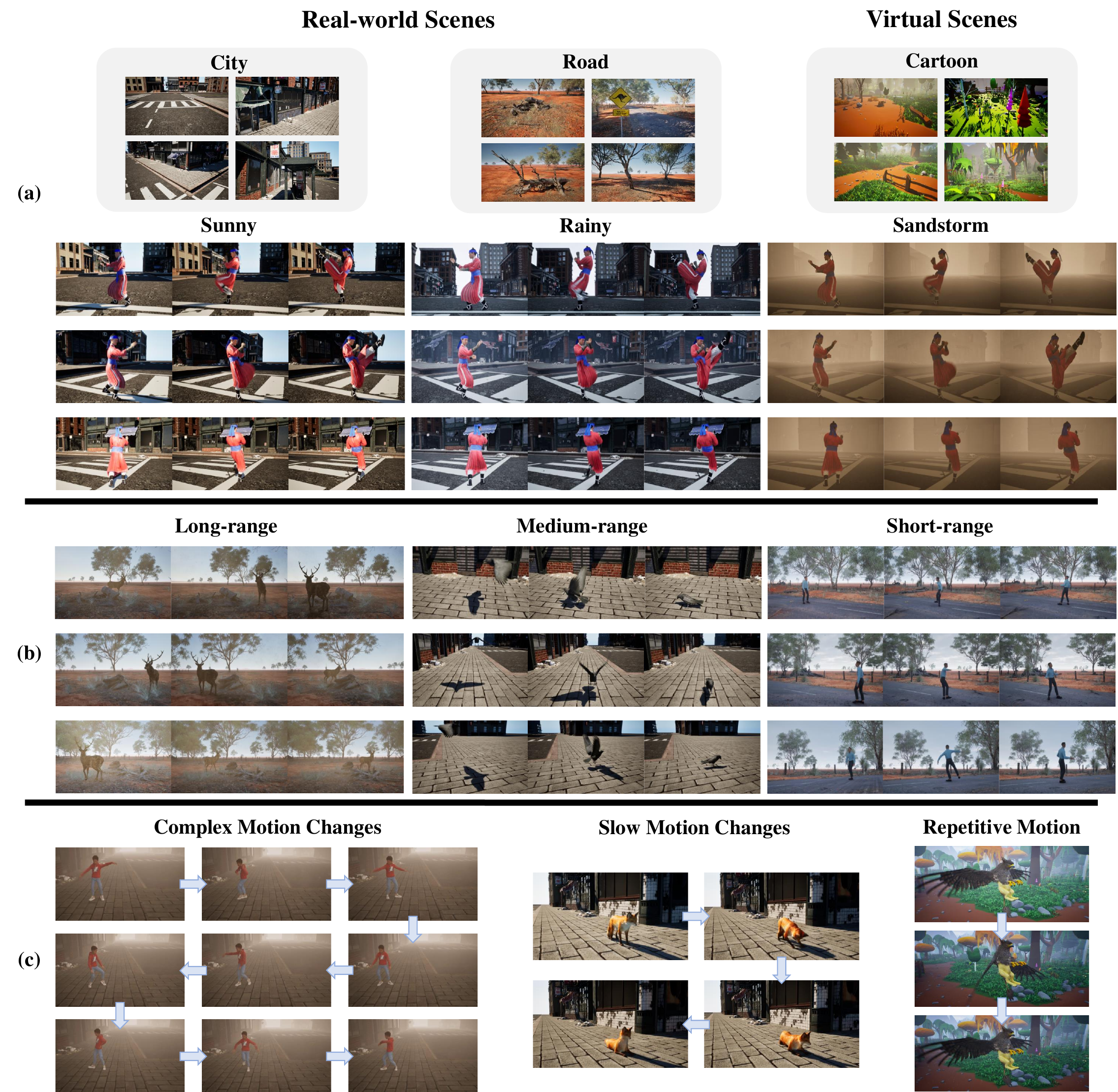}
    \caption{Exhibition of Testing Examples in \dataset.}
    \label{fig:data_category}
    \vspace{-4mm}
\end{figure*}

\begin{figure}
    \centering
    %\fbox{\rule{0pt}{5in} \rule{0.95\linewidth}{0pt}}
    \includegraphics[width=\linewidth]{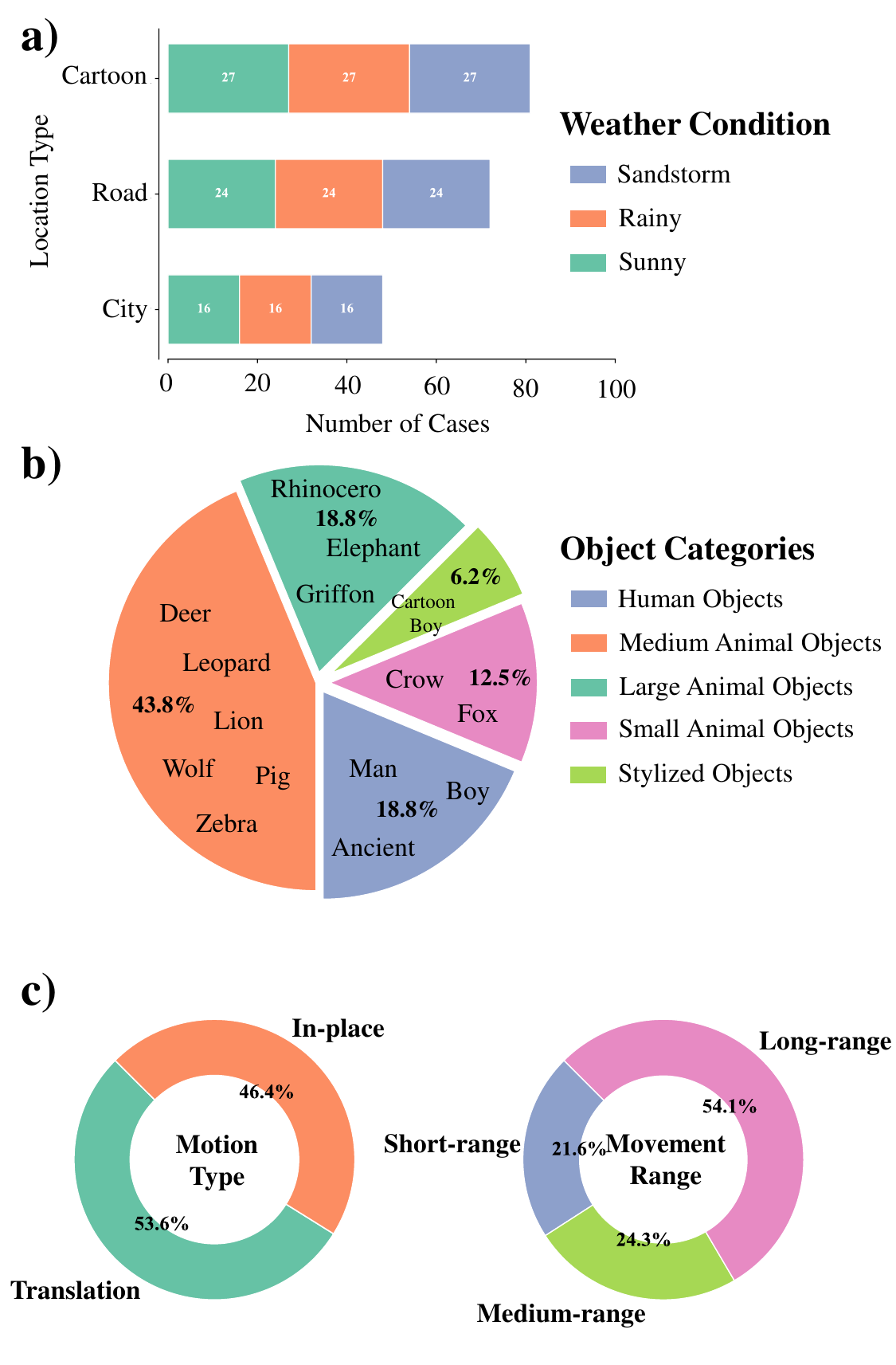}
    \caption{Statistical Distribution of \dataset.}
    \label{fig:data_stat}
    \vspace{-3mm}
\end{figure}

\subsection{4D Dataset}
4D dataset refers to a three-dimensional dataset that includes the time dimension, typically represented by a sequence of videos captured from different viewpoints simultaneously, used to capture the continuous spatial and temporal changes of dynamic scenes or objects. Some early works \cite{lin2022efficient, li2022neural, xu20244k4d} introduced 4D reconstruction datasets for real-world scenarios. However, due to constraints in shooting conditions and limited environments, these datasets lack scene diversity, and the subjects do not exhibit wide-range movement. The 3D dataset, Objaverse dataset \cite{deitke2023objaverse}, also contains 4D data. While the dataset is large, it primarily includes cartoon characters or animals performing localized actions. More recent works, such as DimensionX \cite{sun2024dimensionx} and 4Real \cite{yu20244real}, have proposed methods for generating higher-quality 4D data, yet generating from 360° viewpoints still fails to ensure high fidelity in every view. To address this issue, we construct a richly contextual 4D dataset, \dataset, using the Unreal Engine, which includes objects with significant foreground movement across 360° viewpoints.
% \vspace{-1mm}

\subsection{4D Generation Methods}
With the maturity of 3D generation technologies such as NeRF \cite{mildenhall2021nerf, chan2021pi, muller2022instant, niemeyer2021giraffe, pooledreamfusion, li2024uv, zeng2023ipdreamer, wang2023prolificdreamer} and 3D Gaussian Splatting (3DGS) \cite{kerbl20233d, chung2023luciddreamer, yang2024semantic, liang2024luciddreamer, tang2024dreamgaussian, huang20242d}, 4D generation models based on these 3D generation models have been proposed. A 4D generation model can be defined as a 3D model with temporal information. The initial 4D reconstruction models based on NeRF \cite{pumarola2021d, lin2022efficient, li2022neural, xu20244k4d} collect 4D reconstruction benchmarks and proposed feasible 4D reconstruction methods, with 4K4D \cite{xu20244k4d} ensuring very high-quality 4D reconstruction. Subsequently, due to the higher efficiency and greater flexibility of the 3DGS model compared to NeRF, a 4D reconstruction model based on 3DGS \cite{wu20244d, yang2024real, xu2024representing, li2024st} is introduced. These methods improve generation efficiency while achieving high-quality 4D scene reconstruction on existing 4D reconstruction benchmarks. However, due to limitations in the test data and the deformation field capabilities used in 4D reconstruction methods, existing 4D reconstruction methods struggle to achieve high-quality reconstruction results when dealing with 4D scene reconstruction involving wide-range movement of foreground objects. There are also methods for 4D scene generation conditioned on text or monocular video \cite{wu2024sc4d, xu2024comp4d, chu2024dreamscene4d, zeng2024trans4d}. These methods leverage segmentation models, large language models, and tracking networks to synthesize 4D scenes with significant foreground object movement. However, the quality of the generated 4D scenes is often low and lacks realism. In this paper, we propose the 4D reconstruction method \method, which divides the 4D reconstruction process into two stages: high-quality 3D reconstruction and 4D dynamic progressive fitting. This method provides a more stable and rational construction of high-quality 4D scenes for 4D scene reconstruction tasks involving wide-range spatial movements.

\section{New 4D Benchmark - \dataset}
\label{sec:test_data}

\begin{figure}
    \centering
    \includegraphics[width=0.92\linewidth]{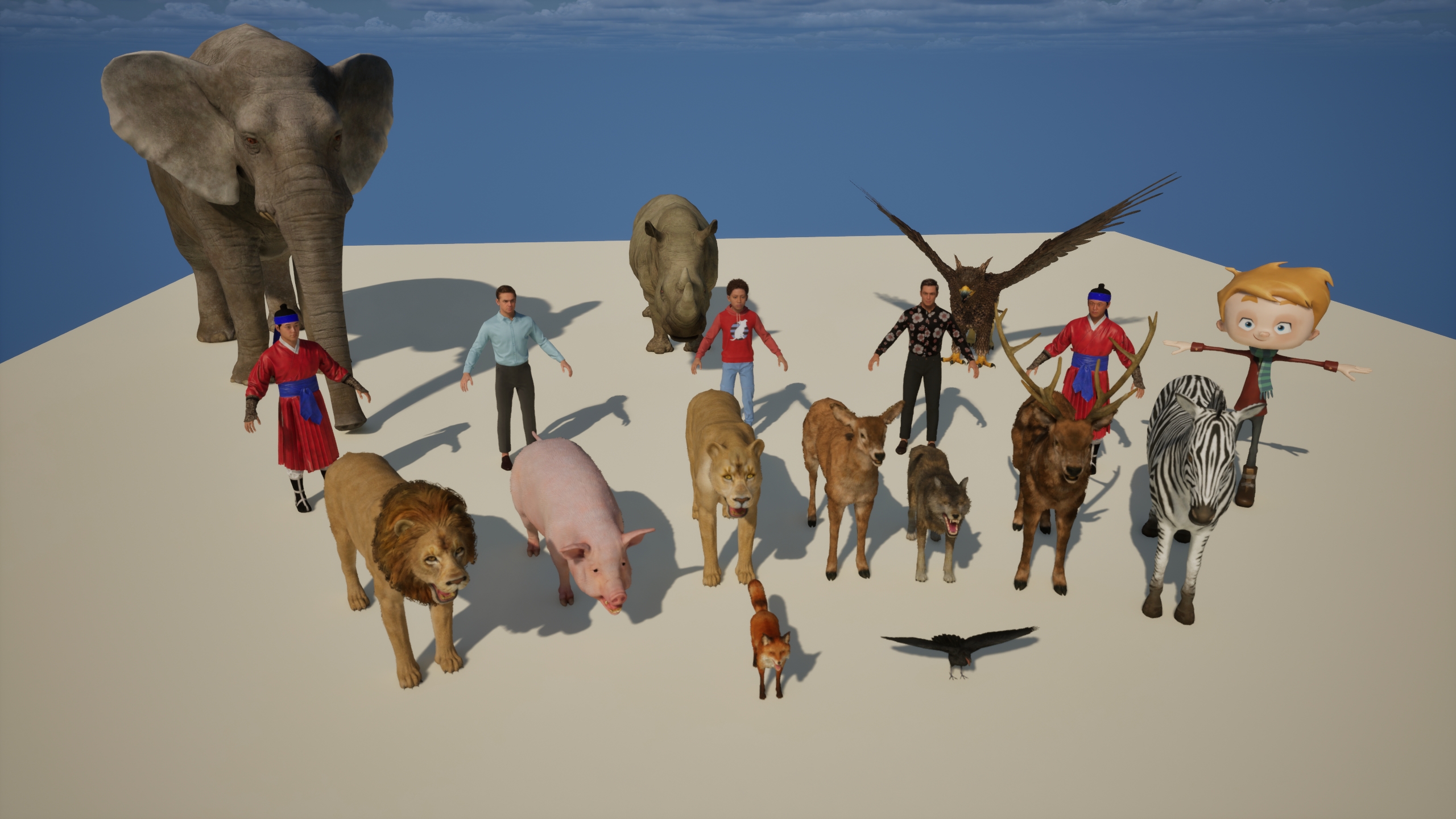}
    \caption{Exhibition of foreground objects}
    \label{fig:dataset_foregrounds}
    \vspace{-3mm}
\end{figure}

\begin{figure*}
    \centering
    \includegraphics[width=\linewidth]{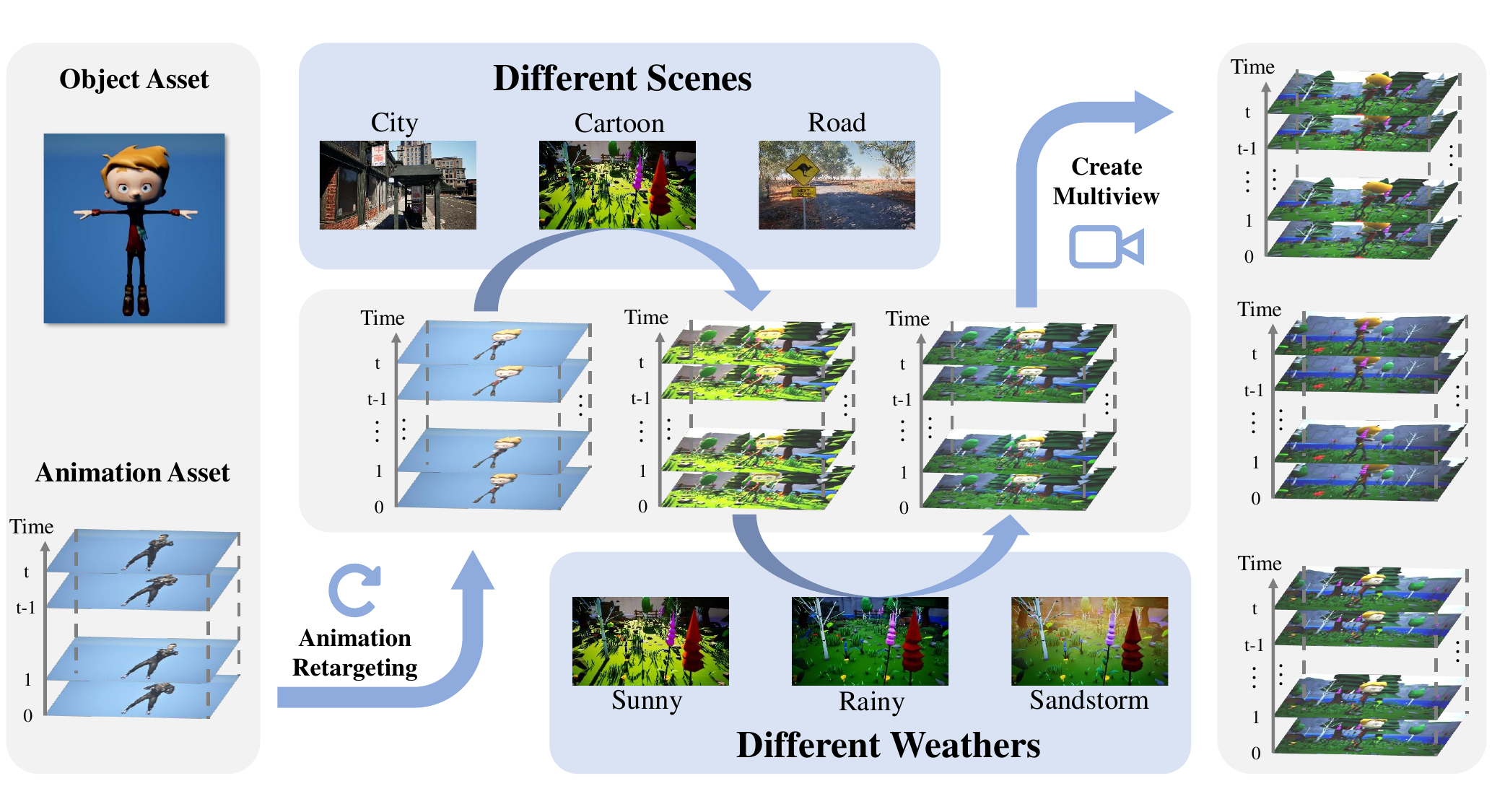}
    \caption{\dataset Construction Pipeline.}
    \label{fig:data_construct}
\end{figure*}

To address the limitation of existing 4D datasets, which lack examples of significant foreground object movement, in this section, we introduce our proposed \dataset, which serves as a new comprehensive benchmark for 4D reconstruction tasks. Next, we will categorize the dataset from various perspectives, count the elements used in the dataset (including foreground objects, weather, actions, etc.), and explain how we curate our \dataset to showcase its comprehensiveness and richness.

\subsection{Data Categories}
To ensure the comprehensiveness and richness of \dataset, we construct the dataset by considering multiple dimensions to capture a wide range of scenarios. \dataset can be categorized into several types in each perspective. First, we consider Scene Types, which include Real-world Scenes, Virtual Scenes, and various weather conditions. These diverse scenes are used to validate the generalization capability of 4D reconstruction methods. In Fig.~\ref{fig:data_category}(a), we present data samples that illustrate environments across different categories. Next, we examine movement distance and speed, which can be broadly classified into short-range, medium-range, and long-range, with different movement trajectories. Longer movement distances and more complex movement paths generally lead to higher reconstruction difficulty. Compared to existing 4D reconstruction datasets, \dataset significantly increases the reconstruction challenge, providing a better benchmark for evaluating the reconstruction ability and stability of 4D reconstruction methods. Examples of movement paths with varying degrees of complexity are shown in Fig.~\ref{fig:data_category}(b). Finally, we consider motion complexity, which encompasses rapid, complex motion changes, slow motion changes, and repetitive motion patterns. This range of action complexities allows for more effective testing of the realism and quality of 4D scene reconstruction methods. We illustrate different levels of action complexity in the 4D data presented in Fig.~\ref{fig:data_category}(c).

\subsection{Statistical Analysis}
In order to present our \dataset more clearly, we conduct statistical analysis on the elements included in the data samples. First, regarding the environment, the scenes we use include cities, cartoon prairies, and country roads. These environments encompass real-world and virtual scenarios, and feature varying amounts of background objects, from densely packed structures, to open and unobstructed landscapes. We also use three types of weather: sunny, rainy, and sandstorm. The sunny weather has minimal background changes, the rainy weather introduces fine raindrops, increasing the difficulty of detail reconstruction, and the sandstorm significantly affects background changes, further increasing the difficulty of 4D reconstruction. The statistics of scenes and weather are shown in Fig.~\ref{fig:data_stat}(a). Next, for Foreground Objects, to ensure the richness and diversity of the data, as shown in Fig.~\ref{fig:dataset_foregrounds}, we selected a variety of foreground objects with different sizes and styles to construct the dataset. The statistics of the Foreground Objects category in the dataset are shown in Fig.~\ref{fig:data_stat}(b). Finally, we perform statistical analysis on motion types and movement ranges, as shown in Fig.~\ref{fig:data_stat}(c). In our \dataset, more challenging motion types and wider movement ranges account for a higher proportion, making \dataset not only comprehensive but also more challenging than previous 4D reconstruction benchmarks.

\subsection{Data Acquisition and Annotation}
We employ several workers with experience in Unreal Engine to build the dataset using open-source 3D assets. For the 4D reconstruction data, we utilize SfM \cite{schoenberger2016sfm} to initialize the scenes. When the quality of 3D scene initialization is poor, we perform further refinement. Additionally, we conduct a user study to evaluate the realism of the motions and the plausibility of the movement paths in the data samples of \dataset. This process helps to filter out unreasonable data, ensuring that our \dataset can serve as a more reliable and validated benchmark for the 4D generation field.

As shown in Fig.~\ref{fig:data_construct}. The dataset \dataset is constructed through a meticulously designed pipeline emphasizing multi-source asset integration, biomechanically diverse motion synthesis, and environmentally dynamic scene composition. This process ensures the generation of a large-scale benchmark tailored for evaluating 4D reconstruction robustness under complex scenarios.

\section{New Baseline Method}
\label{sec:method}

\begin{figure}
    \centering
    \includegraphics[width=\linewidth]{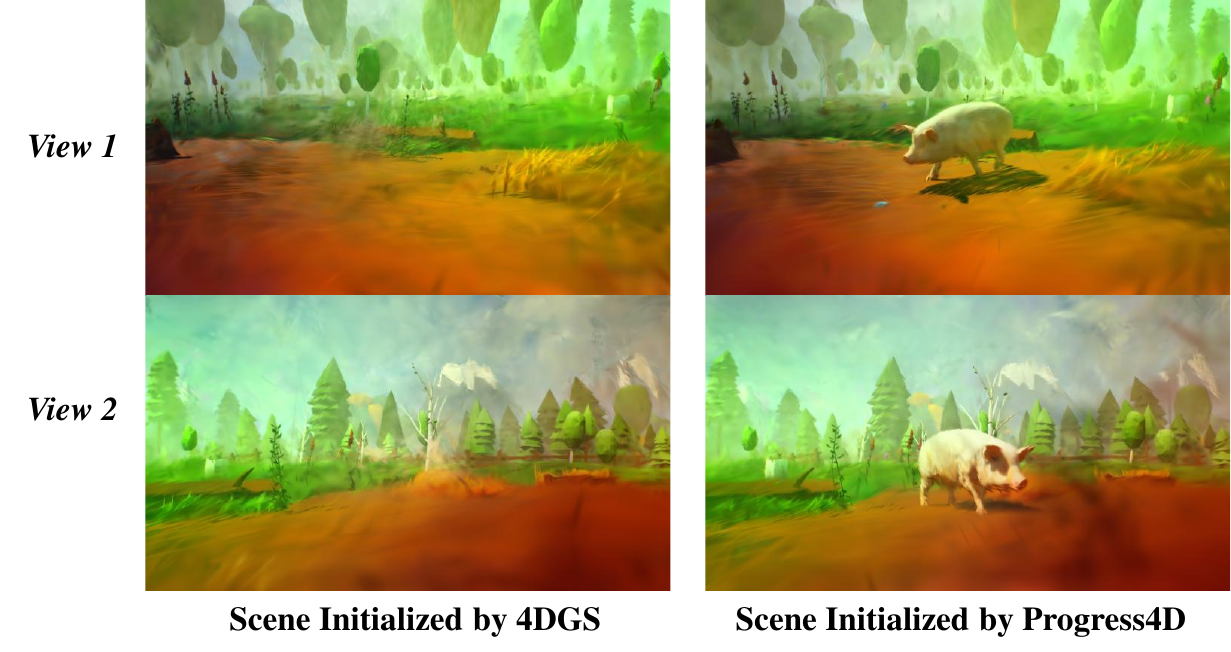}
    \caption{4D scene initial comparison. The reason why existing methods fail to generate high-quality 4D scenes is that when a 4D scene involves wide-range variations, the initialized foreground objects become so blurred that they seem to vanish completely, making it difficult to achieve high-quality 4D synthesis in subsequent generation processes.}
    \label{fig:scene_initial}
    \vspace{-3mm}
\end{figure}

\begin{figure}
    \centering
    \includegraphics[width=\linewidth]{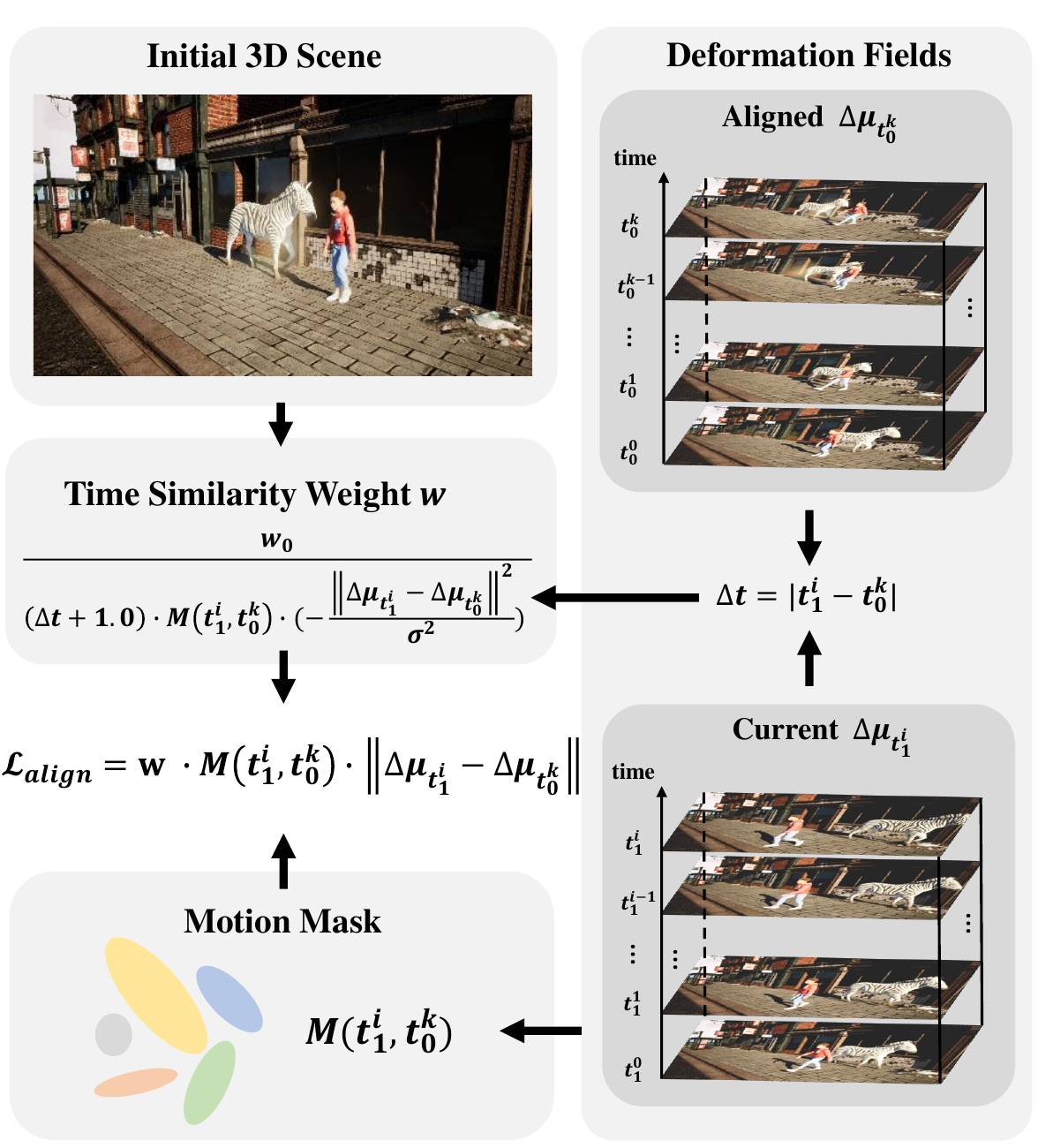}
    %\fbox{\rule{0pt}{5in} \rule{0.95\linewidth}{0pt}}
    \caption{Framework of our \method.}
    \label{fig:method_frame}
    \vspace{-3mm}
\end{figure}

In this section, we detail the process of reconstructing rational 4D scenes with wide-range spatial movement. Firstly, we introduce the preliminaries related to 4D generation. Then, we present the specific algorithm of our \method and analyze how it achieves more stable and higher-quality 4D scene synthesis compared to existing methods.

\subsection{Preliminaries}
Before analyzing 4D generation, it is essential to first understand 3D generation models. Among existing 3D generation models, the 3DGS model is the most flexible and effective. This model consists of multiple 3D Gaussian Points $G$, and each point is represented by a center position $\mu$, covariance $\Sigma$, opacity $\alpha$, and color $c$. The covariance $\Sigma$ is computed using the scaling matrix $S$ and rotation matrix $R$ as $\Sigma = RSS^\top R^\top$. During the rendering process, each pixel of the rendering result is computed by combining $N$ points. This process can be expressed as:
\begin{equation}
G(x) = \sum_{i=1}^{N} \alpha_i \cdot c_i \cdot \exp\left(-\frac{1}{2}(x - \mu_i)^\top \Sigma_i^{-1} (x - \mu_i)\right),
\end{equation}
where $x$ represents the position of a random point in the 3D scene during the rendering process. The 4D model extends the 3D model by incorporating temporal information. The method involves calculating the deformation of each Gaussian Point at time $t$, specifically by introducing a deformation field represented by Multi-Layer Perceptrons (MLPs). This deformation includes changes to the position $\Delta \mu$, scaling $\Delta S$, and rotation $\Delta R$. The deformed attributes of the Gaussian Point $( \mu', R', S' )$ can be represented as:
\begin{equation}
( \mu', R', S' ) = ( \mu + \Delta \mu, R + \Delta R, S + \Delta S).
\end{equation}
Since it is challenging for MLPs to estimate rational wide-range spatial movement directly, existing 4D reconstruction methods struggle to handle 4D scenes with wide-range variations. Next, we explain how our proposed \method achieves stable and high-quality 4D reconstruction in complex scenarios.

\subsection{The \method Framework}
\label{subsec:framework}
To achieve high-quality 4D scene reconstruction, we divide the process into two steps: generating a high-quality 3D scene and gradually fitting the dynamic 4D scene. 

\paragraph{High-quality scene initialization.}
The first step is to create the high-quality 3D scene initialization. Since the test cases in existing 4D reconstruction benchmarks typically involve relatively simple in-place movements, it is easier to initialize a higher-quality 3D scene. However, for 4D scenes with wide-range movement, as shown in Fig.~\ref{fig:scene_initial}, existing 4D reconstruction methods struggle to initialize a high-quality 3D scene. This leads to difficulties in generating high-quality 4D results in complex 4D reconstruction tasks. Therefore, in our method, when initializing the 3D scene, we perform high-quality 3D reconstruction of all objects in the 4D scene in their stationary state, ensuring high-quality 4D scene reconstruction.

\begin{algorithm}[h]
\caption{Optimized Training Pipeline of \method}
\label{app:method}
\begin{algorithmic}[1]
\Require Multi-view video frames $I$, initial 3D Gaussian representation $G^0$, timestep partition $T = {T_0, T_1, T_2}$
\Ensure Optimized 4D scene representation $G^*$
\State \textbf{Stage 1: High-Fidelity 3D Scene Initialization}
\State Optimize $G^0$ with multi-view consistency to minimize $\mathcal{L}_{init}$
\State Freeze the stabilized 3D scene representation
\State \textbf{Stage 2: Progressive Fitting of 4D Dynamics}
\While{Training has not converged}
\For{each timestep $t^i_1 \in T_1$}
\State Compute deformation parameters $(\Delta \mu_{t^i_1}, \Delta S_{t^i_1}, \Delta R_{t^i_1})$ using MLP
\State Identify closest aligned timestep $t^k_0 \in T_0$ based on temporal similarity
\State Compute temporal distance $d_t = |t^i_1 - t^k_0|$
\State Compute adaptive alignment weight:
\begin{equation*}
w_{t^i_1} = \frac{w_0}{d_t + 1.0} \cdot \frac{1}{1 + \exp(-\lVert \Delta \mu_{t^i_1} - \Delta \mu_{t^k_0} \rVert)}
\end{equation*}
\State Compute timestep alignment loss:
\begin{equation*}
\mathcal{L}_{align}(t^i_1) = w_{t^i_1} \cdot \mathbb{I} ( \lVert \Delta \mu_{t^i_1} - \Delta \mu_{t^k_0} \rVert > \tau ) \cdot \lVert \Delta \mu_{t^i_1} - \Delta \mu_{t^k_0} \rVert
\end{equation*}
\EndFor
\State Compute overall loss:
\begin{equation*}
\mathcal{L} = \mathcal{L}_1 + \mathcal{L}_{tv} + \sum_{t^i_1 \in T_1} \mathcal{L}_{align}(t^i_1)
\end{equation*}
\State Update timestep sets:
\State \hspace{1em} Move aligned timesteps $T_1 \rightarrow T_0$
\State \hspace{1em} Select nearest timesteps from $T_2 \rightarrow T_1$
\EndWhile
\State \Return optimized 4D scene representation $G^*$
\end{algorithmic}
\end{algorithm}

\paragraph{4D dynamics progressive fitting.}
With the high-quality initialized 3D scene, the next step is implementing the dynamics in the complex 4D scene. Implementing rational wide-range deformation is difficult for a randomly initialized deformation field. We adopt a step-by-step data loading strategy to align the 4D scene with the initialized 3D scene and the inference multi-view video. This alignment is done progressively based on the similarity between the 4D scene and each frame of the multi-view video, from high similarity to low similarity. Specifically, we divide the timesteps of the 4D scene into three parts: the aligned timesteps $T_{0} = \{ t^{0}_{0}, t^{1}_{0}, ... \}$, the timesteps currently being aligned $T_{1} = \{ t^{0}_{1}, t^{1}_{1}, ... \}$, and the timesteps yet to be aligned $ T_{2} = \{ t^{0}_{2}, t^{1}_{2}, ... \} $. During training, the 4D scene is mainly aligned with the video frames $\hat{I}$ corresponding to the timesteps in $T_{1}$. The data loading update strategy is to add all the aligned timesteps in $T_{1}$ to $T_{0}$, and add the timesteps in $T_{2}$ that are closest to the previous $T_{1}$ timesteps into $T_{1}$. This gradual process allows the deformation field to adapt to the dynamics of the 4D scene, enabling it to achieve complex wide-range movement.

To ensure the training of the 4D scene during the timesteps in $T_{1}$ is as stable as possible, we introduce an advanced timestep alignment loss $\mathcal{L}_{align}$, intricately guided by a motion mask $M(t^{i}_{1}, t^{k}_{0})$ derived from per-frame kinematic saliency. This mask prioritizes regions of significant dynamic variation, enhancing the deformation field’s precision in capturing complex spatial-temporal transitions. Let $\Delta \mu_{t^{i}_{1}}$ represent the position deformation estimated by the deformation field at a particular timestep $t^{i}_{1}$ in $T_{1}$, and let $t^{k}_{0}$ denote the timestep in $T_{0}$ most similar to $t^{i}_{1}$, then $\mathcal{L}_{align}$ is expressed with a richly parameterized weight as:
\begin{equation}
\begin{aligned}
    w &= \frac{w_0}{|t^{i}_{1} - t^{k}_{0}| + 1.0} \cdot \frac{1}{1 + \exp\left(-\lVert \Delta \mu_{t^{i}_{1}} - \Delta \mu_{t^{k}_{0}} \rVert\right)}, \\
    \mathcal{L}_{align} &= w \cdot \underbrace{\mathbb{I}\left(\lVert \Delta \mu_{t^{i}_{1}} - \Delta \mu_{t^{k}_{0}} \rVert > \tau\right)}_{M(t^{i}_{1}, t^{k}_{0})} \cdot \lVert \Delta \mu_{t^{i}_{1}} - \Delta \mu_{t^{k}_{0}} \rVert,
\end{aligned}
\end{equation}
where $w_{0}$ is a baseline weight, $\mathbb{I}(\cdot)$ is the indicator function, $\tau$ is a predefined threshold, $M(t^{i}_{1}, t^{k}_{0})$ is the motion mask that emphasizes regions with significant dynamic variation. Similarly to other reconstruction methods \cite{pumarola2021d, fang2022fast, sun2022direct, cao2023hexplane, fridovich2023k, kerbl20233d, wu20244d, xu20244k4d}, we also use $\mathcal{L}_{1}$ color loss and grid-based total-variation loss $\mathcal{L}_{tv}$. The full loss function can be defined as:
\begin{equation}
\mathcal{L} = \mathcal{L}_{1} + \mathcal{L}_{tv} + \mathcal{L}_{align}.
\end{equation}
To comprehensively illustrate the optimization process of \method, we provide a specific process in Fig.~\ref{fig:method_frame} and a detailed explanation of its algorithmic workflow~\ref{app:method} for reconstructing 4D scenes with large-scale spatial movement. The algorithm of \method consists of two key stages: (1) high-fidelity 3D scene initialization and (2) progressive adaptation to 4D dynamics. 

\section{Experiment}
\label{sec:exp}

\begin{table}[t]
\centering
\caption{Quantitative Comparison of 4D Reconstruction Methods (Best in Bold)}
\label{tab:quantitative_comparison}
\begin{tabular}{@{}lcccc@{}}
\toprule
\textbf{Method} & \textbf{L1$\downarrow$} & \textbf{PSNR$\uparrow$} & \textbf{SSIM$\uparrow$} & \textbf{LPIPS$\downarrow$} \\
\midrule
% \multicolumn{5}{c}{\textit{Monocular Video to 4D Reconstruction methods}} \\
Dreamscene4D & 0.0168 & 21.33 & 0.75 & 0.30 \\
SC4D & 0.0165 & 21.72 & 0.77 & 0.29 \\ \hline
% \addlinespace
% \multicolumn{5}{c}{\textit{4D Reconstruction methods}} \\
4DGS & 0.0155 & 24.65 & 0.82 & 0.25 \\
ST-4DGS & 0.0153 & 26.35 & 0.84 & 0.24 \\ \hline
% \rowcolor{blue!10}
\textbf{Ours} & \textbf{0.0145} & \textbf{28.86} & \textbf{0.87} & \textbf{0.22} \\
\bottomrule
\end{tabular}
\vspace{0.2cm}
\end{table}

\begin{figure*}
    \centering
    % \fbox{\rule{0pt}{6.5in} \rule{0.95\linewidth}{0pt}}
    \includegraphics[width=\linewidth]{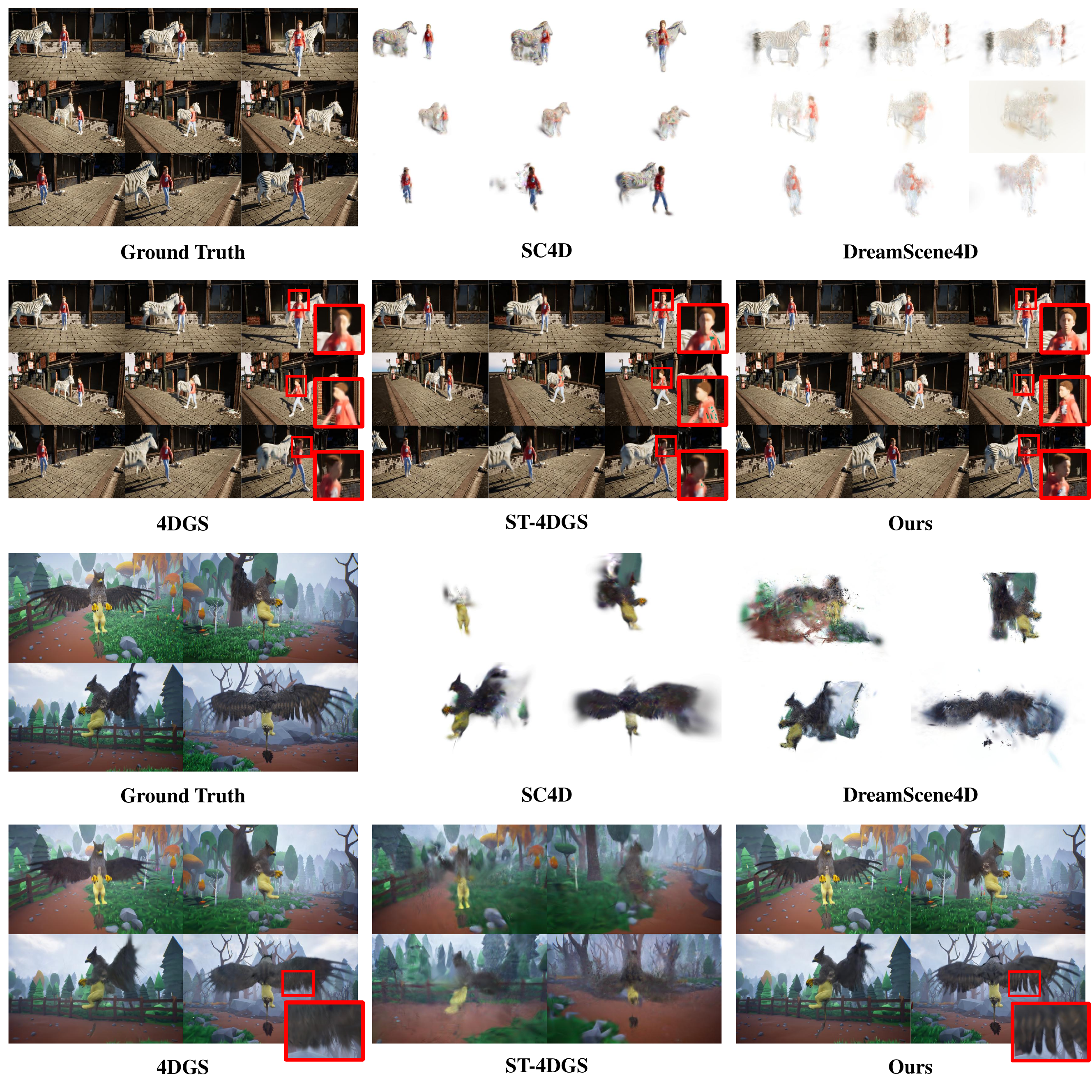}
    \caption{Qualitative comparison. For images that seem similar, we zoom in on the same regions in both our outputs and the comparison images to highlight the finer details of our \method.}
    \label{fig:qual_comp}
\end{figure*}

\begin{figure}
    \centering
    % \fbox{\rule{0pt}{2in} \rule{0.95\linewidth}{0pt}}
    \includegraphics[width=0.95\linewidth]{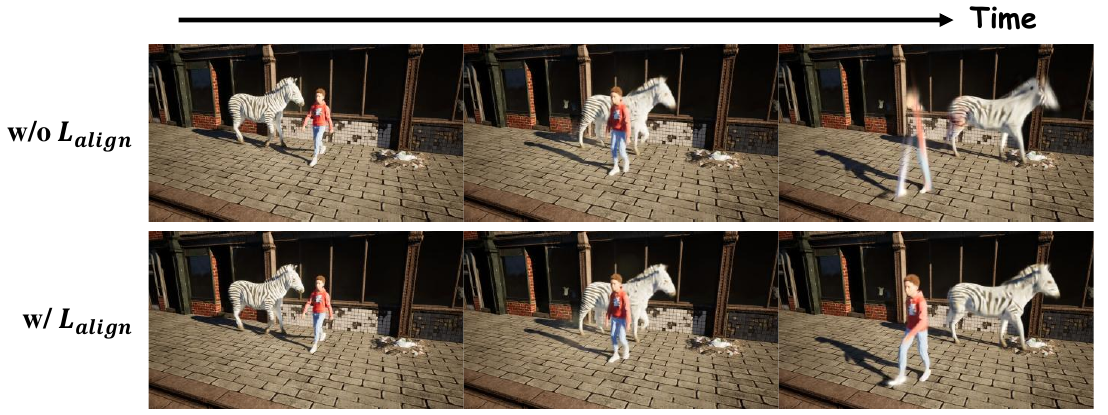}
    \caption{Ablation study of timestep alignment loss.}
    \label{fig:ablation}
    \vspace{-2mm}
\end{figure}

\subsection{Implementation Details}

\paragraph{Data Setting.}
Consistent with previous 4D benchmarks, \dataset also provides sequential frames from multiple viewpoints. The input data is processed into specific dataset formats as each method requires. Each testing case includes 40 viewpoints, with each viewpoint containing 60 to 150 frames. Most testing cases involve wide-range spatial movement.
\vspace{-3mm}

\paragraph{Optimization.}
All experiments are conducted on 8 RTX 4090 GPUs, and each 4D reconstruction task is completed on a single RTX 4090 GPU. We utilize the Adam optimizer \cite{Kingma2014adam} with a learning rate of $1.6e^{-4}$ to optimize our 4D model. For our \method, the timestep lists $T_{0}$, $T_{1}$, and $T_{2}$ are updated every 1000 iterations.
\vspace{-3mm}

\paragraph{Metrics.}
We employ L1, PSNR, SSIM \cite{Wang2004ssim}, and LPIPS \cite{Zhang2018lpips} to evaluate the performance of 4D generation methods. L1 and PSNR measure the difference between the generated image and its ground truth, where a lower L1 value and a higher PSNR value indicate better reconstruction quality. SSIM measures the structural similarity between the generated image and its ground truth, with higher values indicating better reconstruction quality. LPIPS assesses the perceptual quality of the generated images, with lower values indicating better generation results.
\vspace{-3mm}

\paragraph{Baseline.}
For baseline methods, we compare our \method with other 4D reconstruction methods, including 4DGS \cite{wu20244d} and ST-4DGS \cite{li2024st}, as well as monocular video-to-4D generation methods such as DreamScene4D \cite{chu2024dreamscene4d} and SC4D \cite{wu2024sc4d}, which use segment models and track models \cite{cheng2022xmem, ke2023segment, chu2024zero, ren2024grounded} to achieve rational 4D scene generation. For monocular video-to-4D generation, to ensure fairness in comparison, we only compare the 4D reconstruction results of these methods with the ground truth under the corresponding viewpoint of the input video. To reduce the impact of the background, we additionally introduce a mask for monocular video-to-4D generation methods. All comparisons are based on the official code of the baseline methods available on GitHub.

\subsection{Main Results}
We conduct both qualitative and quantitative comparisons between our \method and other baseline methods on \dataset to validate the effectiveness of \method, while also highlighting the significant challenges that \dataset presents to existing 4D generation methods.

In Table~\ref{tab:quantitative_comparison}, we present the results of the quantitative comparison. It is evident that, for the metrics measuring reconstruction consistency (L1 and PSNR), the metric assessing structural similarity (SSIM), and LPIPS, which gauges the authenticity of generated images, our \method achieves the best performance, thereby demonstrating the effectiveness of proposed 4D reconstruction method \method. At the same time, the testing results of the baseline methods on \dataset are significantly worse than those on other 4D reconstruction benchmarks, proving both the difficulty of our \dataset and the richness of its data samples, while also indicating the stability of our \method.

Additionally, we provide a qualitative comparison in Fig.~\ref{fig:qual_comp}, which visualizes the generated results of our \method and the baseline methods, offering a more intuitive demonstration of the generation quality of our \method and the existing methods. In Sample 1, we showcase the results generated by various methods from different viewpoints and times. For the 4D reconstruction methods 4DGS and ST-4DGS, especially 4DGS, although the background quality is high, the moving objects in the foreground are highly blurred. Additionally, using the default parameter settings in the official code of ST-4DGS fails to produce high-quality 4D scenes, demonstrating that reconstructing 4D scenes with wide-range spatial movement is challenging for existing 4D reconstruction methods. Furthermore, for SC4D and DreamScene4D, which utilize segmentation and tracking models to achieve rational 4D scene generation, it is difficult to achieve clear and accurate 4D generation using the data samples in our \dataset, further proving the challenge of testing 4D scene examples in \dataset. Our method successfully and stably reconstructs rational, high-quality 4D scenes. To further demonstrate the quality of 4D scene generation by our \method, in Sample 2, we show a 4D scene of actions performed in place. Compared to previous approaches, our method's reconstruction results are the clearest, with the best detail preservation.

\subsection{Ablation Study}
In Section~\ref{subsec:framework}, we have already analyzed the contribution of high-quality scene initialization in 4D scene generation. In this section, we analyze the effectiveness of the timestep alignment loss $\mathcal{L}_{align}$ in 4D dynamics progressive fitting. As shown in Fig.~\ref{fig:ablation}, without $\mathcal{L}_{align}$, the scene generation quality is high when the deviation from the initial scene is small. However, as time progresses and the movement distance increases, the foreground significantly deteriorates. With $\mathcal{L}_{align}$, even with substantial deviations, the generation effect remains impressive, proving that the timestep alignment loss plays a crucial role in the dynamic synthesis of 4D scenes.

\section{Conclusion}
\label{sec:conclusion}

In this work, we propose a new 4D reconstruction benchmark, \dataset, which addresses a critical limitation in existing 4D benchmarks, the lack of data with wide-range movements, and fills this gap by introducing data samples that encompass these dynamics. Our \dataset contains diverse and rich content, enabling a more comprehensive evaluation of 4D generation methods. Additionally, to overcome the limitations of previous 4D reconstruction methods in handling wide-range variations, we introduce \method. Our method decomposes 4D scene generation into high-quality 3D scene initialization and progressive 4D dynamic fitting. Both qualitative and quantitative experiments demonstrate that our method generates higher-quality 4D scenes more stably compared to baseline 4D generation methods.

{
    \small
    \bibliographystyle{ieeenat_fullname}
    \bibliography{main}
}

\newpage
\maketitlesupplementary

\section{\dataset Construction Pipeline}
%In this section, we provide a specific construction pipeline of our \dataset...
In this section, we provide a specific construction pipeline of our \dataset.

\subsection{Asset Integration and Motion Synthesis}
The construction of \dataset begins with the integration of heterogeneous 3D assets and biomechanically valid motion adaptation. Foreground objects, including humans, animals, and stylized characters, are sourced from three platforms: rigged humanoid models from Mixamo and Unreal Engine’s MetaHuman, and some other assets (e.g., animal, ancient) from the Unreal Marketplace (FAB). These models are categorized by geometric scale (Small, Medium, Large) and artistic style (realistic, cartoonish) to ensure diversity in object geometry, texture, and semantic representation. Animation assets, which span locomotion and gestures are extracted from Mixamo’s animation library and retargeted to heterogeneous skeletons using Unreal Engine’s Control Rig system. Key technical steps include resolving joint discrepancies through inverse kinematics (IK) solvers to ensure anatomical plausibility and dynamically scaling root motion parameters to adapt animations across varying model proportions, such as adjusting stride lengths for quadrupedal animals or gait cycles for bipedal humans. To amplify reconstruction challenges, motion trajectories are procedurally synthesized into three complexity tiers: short-range linear paths, medium-range curved trajectories, and long-range multi-segment paths. These trajectories are generated via spline-based procedural animation, incorporating collision constraints to enforce kinematic validity while maintaining natural movement patterns. The integration of diverse objects, biomechanical motion adaptation, and complex trajectory design collectively ensure that \dataset captures a broad spectrum of motion scenarios critical for evaluating 4D reconstruction robustness.

\begin{figure*}[h!]
 \centering
 \includegraphics[width=\linewidth]{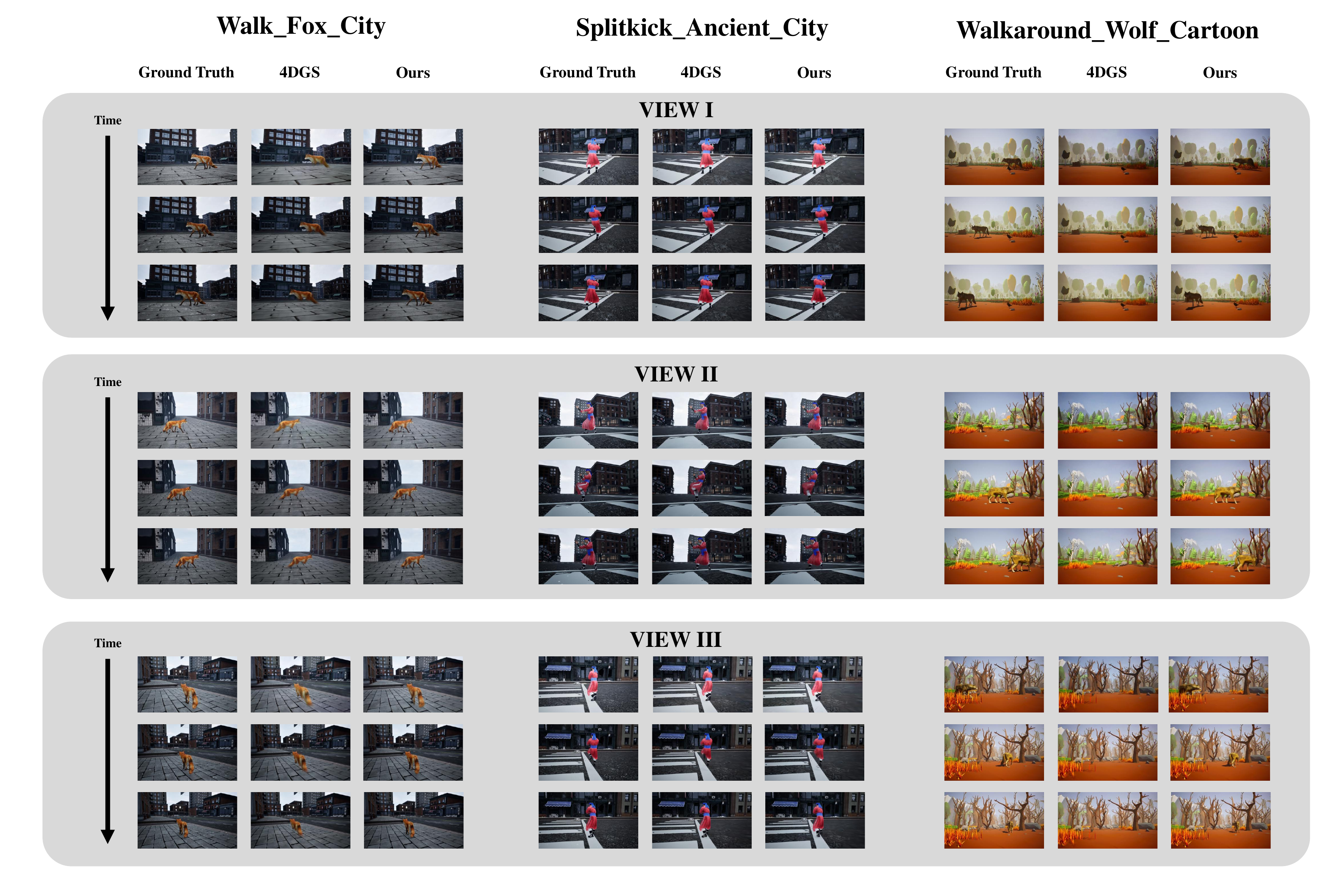}
 \caption{Other Qualitative Comparison Results.}
 \label{fig:add_dataresults}
\end{figure*}

\subsection{Environmentally Dynamic Scene Composition}
Dynamic scene composition in \dataset emphasizes environmental variability and multi-view geometric consistency. Scenes are constructed using modular assets from Unreal Engine’s FAB library, categorized into real-world environments (urban streets, country roads), virtual settings (cartoon prairies), and hybrid configurations blending synthetic and real-world elements. Environmental dynamics are parametrically controlled via the Ultra Dynamic Sky plugin, introducing three weather states: sunny conditions with stable illumination and minimal atmospheric interference, rainy weather with particle-based raindrops, surface wetness effects, and refractive distortions, and sandstorms characterized by volumetric dust clouds, wind-driven debris, and visibility degradation. The multi-view data acquisition is implemented through a configurable array of CineCamera Actors, strategically positioned with parametric control over azimuth (0°–360°), elevation (-15°–15°), and radial distance (2–10 meters) relative to scene centroids. Each camera is configured with a 15.0mm focal length, f/2.8 aperture, and a Universal Zoom lens system paired with a 16:9 Digital Film sensor backplate to ensure cinematic depth-of-field and optical consistency across viewpoints. Scene illumination and environmental conditions are dynamically modulated under three configurations (sunny, rainy, and sandstorm), with synchronized rendering performed across all cameras to generate 60 temporally aligned RGB sequences per scene. Rendered at 2K resolution (2560×1440 pixels) and 30 FPS. This setup ensures high-fidelity multi-perspective capture while maintaining photometric and geometric coherence, enabling robust evaluation of 4D reconstruction algorithms under diverse environmental and viewpoint constraints.

\section{Algorithm Procedure of \method}
Below, we present a structured breakdown of the methodology, elaborating on the critical aspects of optimization and alignment mechanisms involved. 

\subsection{High-Quality 3D Scene Initialization}
The foundation of \method relies on constructing a geometrically precise and visually consistent 3D scene. To this end, the Gaussian representation $G^0$ is optimized by refining its opacity $\alpha$, color attributes $c$, spatial coordinates $\mu$, and covariance matrix $\Sigma$ through multi-view image supervision. The initialization objective function is formulated as:
\begin{equation}
\mathcal{L}_{init} = \mathcal{L}_1 + \mathcal{L}_{tv},
\end{equation}
where $\mathcal{L}_1$ represents a pixel-wise color reconstruction loss, ensuring photometric consistency, while $\mathcal{L}_{tv}$ enforces smoothness regularization to suppress noise and artifacts. Once convergence is achieved, the optimized 3D Gaussian \cite{kerbl20233d} representation $G^*$ is retained as a stable foundation for subsequent temporal adaptation.

\subsection{Progressive Fitting of 4D Dynamics}
Following 3D initialization, the next stage involves learning temporally coherent deformations to accurately model large-scale object and scene transformations. Given the inherent complexity of estimating non-rigid motion, we introduce a progressive timestep alignment mechanism that progressively refines the reconstructed 4D scene. The temporal optimization process partitions timesteps into three categories:
\begin{itemize}
\item $T_0$: Timesteps that have been successfully aligned and stabilized.
\item $T_1$: Timesteps undergoing active alignment and refinement.
\item $T_2$: Future timesteps yet to be processed.
\end{itemize}
At each iteration, deformation parameters $\Delta \mu$, $\Delta S$, and $\Delta R$ are estimated using the deformation field. To ensure robust alignment, a similarity metric is computed between timesteps in $T_1$ and their closest reference frames in $T_0$, defining an adaptive weighting function that modulates the influence of each timestep in the optimization process.

By iteratively refining timestep alignments and leveraging adaptive deformation estimation, \method attains a stable and high-fidelity 4D representation. The integration of spatial smoothness priors, motion-aware losses, and progressive optimization ensures seamless temporal transitions while preserving geometric precision. This approach surpasses conventional methods by significantly enhancing reconstruction stability, robustness, and overall visual plausibility.

\section{Additional Generated Results}
To further demonstrate the efficacy of \method in reconstructing complex 4D scenes, we present additional qualitative comparisons with 4DGS \cite{wu20244d} across various dynamic environments. These comparisons highlight the method’s ability to handle challenging scenarios involving extensive motion, occlusions, and intricate structural deformations.

As shown in Fig.~\ref{fig:add_dataresults}, each example presents a sequence of rendered frames, demonstrating our reconstructions' temporal coherence and spatial accuracy. The results highlight the smoothness of temporal transitions and the preservation of fine-grained details, even in the presence of wide-range spatial movements. By integrating our progressive alignment strategy, artifacts and distortions are minimized, resulting in more visually plausible reconstructions than those produced by existing methods.

The robustness of \method is further validated by its performance across diverse scene types, including both virtual and real-world scenarios. Our method effectively generalizes to varying motion patterns and wide movement ranges, establishing it as a reliable solution for high-fidelity 4D scene reconstruction. Moreover, \method consistently outperforms baseline approaches in terms of rendering clarity, geometric consistency, and overall realism.

\end{document}